\documentclass[11pt]{article}

\usepackage[preprint]{acl}

\usepackage{times}
\usepackage{latexsym}
\usepackage[T1]{fontenc}
\usepackage[utf8]{inputenc}
\usepackage{microtype}
\usepackage{inconsolata}
\usepackage{graphicx}

\usepackage{url}
\usepackage{booktabs}
\usepackage{multirow}
\usepackage{amsmath}
\usepackage{amssymb}
\usepackage{mathtools}
\usepackage{bm}
\usepackage{xcolor}
\usepackage{tikz}
\usetikzlibrary{positioning, arrows.meta, calc}
\usepackage[capitalize,noabbrev]{cleveref}

\newcommand{\auroc}{\mbox{\textsc{auroc}}}

\title{The Knowing-Saying Gap: When Probes See Errors that Confidence Misses}

\author{Jyotin Goel\thanks{Equal contribution.} \quad Ipshita Bandyopadhyay\footnotemark[1] \quad Justin Shenk \\
Correspondence: \texttt{jyotinofficialcc@gmail.com}
}

\begin{document}

\maketitle

\begin{abstract}
Linear probes detect corrupted context in language models with near-perfect
accuracy, yet this does not translate into reliable failure prediction. The
result is a dissociation with direct implications for deployment monitoring.
Across multi-hop arithmetic chains, probes that detect corruption turn out to
be \emph{uninformative} about final answer correctness; models forced into
structured confidence formats collapse to two values with indistinguishable
error rates; and probe persistence across hops fails to separate correct from
incorrect outcomes, refuting our pre-registered ``persistence beats peak''
hypothesis. This pattern of \emph{knowing but not saying} generalises across
model families including reasoning models. As a real-time monitor, probe-based
interventions are sharply model and error-type dependent: branch-and-pick is net-positive across models and uniquely non-breaking on Llama-3.1-8B(4 rescued, 0 broken), while
reprompt and replace-prior break correct traces at roughly the rate they rescue
wrong ones. Probe-based monitoring is a necessary complement to verbalised
confidence, but no single intervention dominates, and the deployable answer is
model-aware, error-type-aware routing.
\end{abstract}

\section{Introduction}

Large language models are increasingly deployed not just to answer questions
but to act: writing and executing code, navigating multi-step plans, and
operating as nodes in agentic pipelines
\citep{wang2024survey,yao2023react,schick2023toolformer}. In these settings, a
wrong answer is not merely unhelpful, it is a seed. An error at step one
propagates into step two's context, until the accumulated deviation is
unrecoverable \citep{huang2023survey,pan2023automatically}. The model continues
in the same fluent register with no hedging, no slowdown, and no visible sign
that anything has gone wrong \citep{kadavath2022know,lin2022teaching}. This
failure mode is already observable in production. Coding agents such as
SWE-agent \citep{yang2024sweagent}, Devin \citep{cognition2024devin}, and Claude
Code \citep{anthropic2024claudecode} operate in edit-execute-observe loops where
a misdiagnosed root cause at turn one constrains every subsequent
patch. SWE-bench evaluations show that even the strongest agents fail on the
majority of tasks \citep{jimenez2024swebench}, and qualitative analysis suggests
that cascading context corruption, rather than capability, is a primary
bottleneck \citep{yang2024sweagent}. The same structure appears in LLM-based
medical decision support \citep{singhal2023large}, long-horizon task planning
\citep{valmeekam2023planning}, and multi-agent coordination
\citep{chan2023chateval}: any setting where a model's output feeds back into its
own future context.

This fluency without fidelity is the central reliability problem of deployed
language models, and it will intensify. As inference-time compute scales and
models are given longer context windows, tool access, and persistent memory, the
gap between what a model has internally computed and what it chooses to say
becomes both larger and more consequential
\citep{anthropic2024claude,openai2024gpt4,wei2022chain}. A model that stutters on
uncertain ground would be easy to monitor. A model that produces wrong answers
in the same confident register as correct ones requires a fundamentally
different approach.

The natural response is to ask the model directly. Elicited confidence,
chain-of-thought self-critique, and structured verification have all been
proposed as mitigations
\citep{kadavath2022know,xiong2024can,weng2023large,shinn2023reflexion}. These
share an implicit assumption: that the model's internal state is accessible
through its verbal outputs. We show that this assumption fails in an informative
way.

\paragraph{This paper.}
We study multi-hop arithmetic chains with silently corrupted prior context, a
controlled setting that isolates the core failure mode of production agentic
loops while holding the surface form of the task fixed. A linear probe on the
residual stream detects the corrupted context with near-perfect accuracy, yet
does not predict whether the final answer is wrong. Structured confidence
elicitation collapses to a binary signal with indistinguishable error rates.
Real-time interventions based on the probe signal vary sharply by model and
error type: branch-and-pick is the only strictly non-breaking policy, while
reprompt and replace-prior rescue wrong traces and break correct ones at
comparable rates.
Together these constitute the \emph{knowing but not saying} dissociation. The
model has linearly encoded the relevant fact about its context, but that
information does not propagate into its verbal outputs or downstream behaviour.
As coding agents and agentic pipelines become standard infrastructure,
probe-based monitoring of internal state is a necessary complement to verbalised
confidence, since the alternative is trusting a channel we show to be
uninformative.

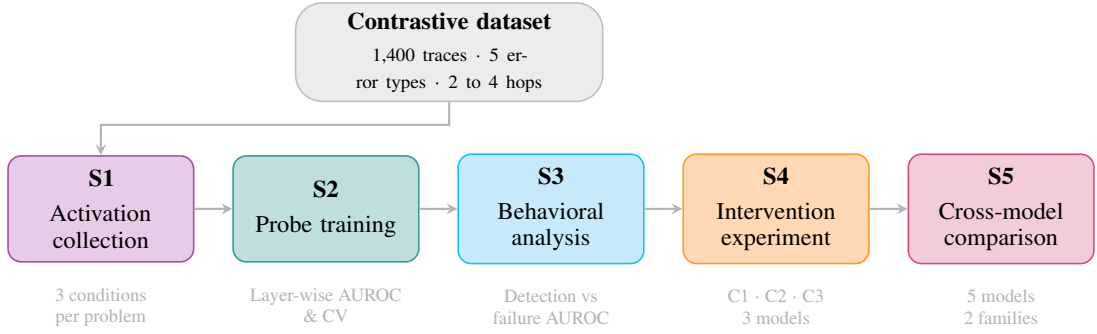
\begin{figure*}[t]
\centering
\begin{tikzpicture}[
  node distance = 0.5cm,
  box/.style = {
    rectangle, rounded corners=5pt,
    minimum width=2.3cm, minimum height=1.4cm,
    text width=2.1cm, align=center,
    font=\small, inner sep=5pt,
    line width=0.6pt
  },
  s1box/.style = {box, fill=violet!20,      draw=violet!70},
  s2box/.style = {box, fill=teal!25,        draw=teal!80},
  s3box/.style = {box, fill=cyan!20,        draw=cyan!70},
  s4box/.style = {box, fill=orange!30,      draw=orange!80},
  s5box/.style = {box, fill=purple!20,      draw=purple!70},
  graybox/.style = {
    rectangle, rounded corners=10pt,
    fill=gray!15, draw=gray!50,
    minimum height=0.9cm, minimum width=4.0cm,
    text width=3.8cm, align=center,
    font=\small, line width=0.6pt},
  arr/.style  = {-stealth, line width=0.8pt, gray!60},
  lbl/.style  = {font=\scriptsize\color{gray!60}, align=center}
]

\node[graybox] (dataset)
  {\textbf{Contrastive dataset}\\[2pt]
   \scriptsize 1,400 traces $\cdot$ 5 error types $\cdot$ 2 to 4 hops};

\node[s1box, below=0.7cm of dataset, xshift=-4.6cm] (s1)
  {\textbf{S1}\\[3pt] Activation collection};

\node[s2box, right=of s1] (s2)
  {\textbf{S2}\\[3pt] Probe training};

\node[s3box, right=of s2] (s3)
  {\textbf{S3}\\[3pt] Behavioral analysis};

\node[s4box, right=of s3] (s4)
  {\textbf{S4}\\[3pt] Intervention experiment};

\node[s5box, right=of s4] (s5)
  {\textbf{S5}\\[3pt] Cross-model comparison};

\node[lbl, below=0.2cm of s1]{3 conditions\\per problem};
\node[lbl, below=0.2cm of s2]{Layer-wise AUROC\\\& CV};
\node[lbl, below=0.2cm of s3]{Detection vs\\failure AUROC};
\node[lbl, below=0.2cm of s4]{C1 $\cdot$ C2 $\cdot$ C3\\3 models};
\node[lbl, below=0.2cm of s5]{5 models\\2 families};

\draw[arr] (dataset.south) -- ++(0,-0.3) -| (s1.north);
\draw[arr] (s1) -- (s2);
\draw[arr] (s2) -- (s3);
\draw[arr] (s3) -- (s4);
\draw[arr] (s4) -- (s5);

\end{tikzpicture}
\caption{Overview of the five-stage experimental pipeline. S1 collects
activations and surface signals; S2 trains a layer-wise linear probe; S3
measures the detection versus failure dissociation; S4 evaluates three
probe-triggered intervention policies; S5 aggregates results across all five
models.}
\label{fig:pipeline}
\end{figure*}

\paragraph{Contributions.}
(1) A contrastive multi-hop math dataset of 1{,}400 traces over 500 base
problems with five error types, evaluated across two model families (Qwen and
Llama) spanning instruction-tuned and reasoning variants (\Cref{sec:dataset}).
(2) A characterisation of the detection versus failure-prediction collapse, with
a head-to-head against six surface uncertainty signals, replicated across models
(\Cref{sec:results}). (3) A controlled comparison of three probe-triggered
interventions with an error-type breakdown and cross-model analysis, motivating
model-aware routed policies (\Cref{sec:intervention_results}). (4) A
pre-registered hypothesis that we report as refuted: persistence of probe
activation across hops does not predict failure(\Cref{app:persistence}). (5) Two robustness controls,
base-problem-grouped cross-validation and leave-one-error-type-out evaluation,
showing the detection signal is neither a fold-leakage artifact nor a per-type
surface signature (\Cref{app:grouped_cv,app:heldout_type}).

\section{Related work}

\paragraph{Probing internal representations.}
Linear probes have been used to extract structural and semantic features from
transformer activations, from syntax \citep{hewitt2019structural} to
truthfulness \citep{burns2023latent,marks2023geometry} and refusal
\citep{arditi2024refusal}. Our finding extends probing to \emph{operational
state}, namely whether an ongoing computation is operating on corrupted input.
We emphasise that successful probing of state does not entail success at
predicting behaviour, a distinction we believe has not been pulled apart this
cleanly before.

\paragraph{Calibration and verbalised confidence.}
A growing literature trains or prompts models to express uncertainty in language
\citep{kadavath2022know,tian2023ask,lin2022teaching,xiong2024can}. Most of this
work treats miscalibration as a noisy quantitative phenomenon, with confidences
that are too high or insufficiently spread. We argue that, at least for the
models and task we study, verbalised confidence is \emph{categorically}
miscalibrated: it functions as a binary plausibility filter on the output rather
than as graded introspection on the chain.

\paragraph{Chain-of-thought faithfulness.}
Recent work has argued that chain-of-thought explanations can be unfaithful
\citep{turpin2023language,lanham2023measuring,paul2024making}. We contribute a
positive \emph{decoding} result. The intermediate state \emph{does} contain the
corruption signal at high fidelity; the unfaithfulness lies in whether that
information is consulted by the surface-output layers that ultimately produce the
answer string.

\paragraph{Process supervision and self-correction.}
Process reward models \citep{lightman2024lets} and self-consistency
\citep{wang2023self} attack the same problem with surface signals. We use an
internal probe as the scoring function for branch-and-pick. Our earlier
unpublished work claimed that this gave a strict asymmetric safety property. The
present paper revises that claim and shows it survives only when conditioned on
error type.

\section{Dataset}
\label{sec:dataset}

We construct a contrastive multi-hop math dataset of 1,400 traces over 500 base
problems, generated deterministically from a single seed (42) with no human
annotation and no language-model involvement (full generation procedure in
\Cref{app:dataset}).

\paragraph{Problem generation.}
Twelve subcategory-specific template functions cover arithmetic, percentages,
rates, fractions, counting, geometry, compound interest, and nested fractions
across 2 to 4 hops. Each draws numeric parameters from fixed candidate sets via a
seeded \texttt{random.Random} instance and computes every intermediate result
exactly, so prompts, correct responses, and ground-truth values are derived
algebraically rather than sampled from a model.

\paragraph{Error injection.}
Each base problem yields a \texttt{clean} variant and one to three
\texttt{error\_at\_k} variants in which the assistant turn at hop $k$ is replaced
by a synthetically wrong response computed deterministically from the correct
operands. Five error types span qualitatively distinct failure modes:
\texttt{off\_by\_one} ($+1$), \texttt{wrong\_operator} (e.g.\ $\times\to+$),
\texttt{wrong\_unit} (correct magnitude, wrong unit), \texttt{magnitude\_error}
($\times 10$), and \texttt{wrong\_percentage\_base} ($100-p$ for $p$); see
\Cref{app:dataset} for motivation. Types are assigned by cycling through the five
in order ($\sim$100 traces each), and all hops downstream of the injection are
re-templated from the wrong upstream value, so propagation is realistic rather
than hand-constructed. Cycling assigns each error type to $\sim$100 of the 500
base problems; because longer problems contribute multiple \texttt{error\_at\_k}
variants, this yields roughly 180 corrupted traces per type across the 900
corrupted traces. \Cref{tab:variants} gives the variant distribution.

\begin{table}[t]
\begin{center}
\small
\begin{tabular}{lc}
\toprule
Variant & Traces \\
\midrule
\texttt{clean}        & 500 \\
\texttt{error\_at\_1} & 500 \\
\texttt{error\_at\_2} & 250 (3-hop and longer only) \\
\texttt{error\_at\_3} & 150 (4-hop only) \\
\midrule
Total & 1,400 \\
\bottomrule
\end{tabular}
\end{center}
\caption{Variant distribution across the dataset.}
\label{tab:variants}
\end{table}

\section{Methodology}
\label{sec:methods}

\subsection{Experimental setup}

We investigate the knowing-saying gap across five models spanning two
architecture families, two scales, and one reasoning variant. All models are
loaded in bf16 full precision with frozen weights throughout, so no fine-tuning
occurs at any stage. The probe is the only learned component, trained
exclusively on activations extracted from the frozen model. Models are
summarised in \Cref{tab:models}.

The thinking-mode variant differs from the standard pipeline in exactly one way:
extended chain-of-thought reasoning is enabled at inference time. All other
hyperparameters, dataset subsets, and evaluation procedures are held fixed,
making this a single-variable ablation of whether chain-of-thought reasoning
closes the knowing-saying gap.

\begin{table}[t]
\begin{center}
\small
\setlength{\tabcolsep}{3pt}
\begin{tabular}{lcccc}
\toprule
Model & Params & Layers & $d_\text{model}$ & Mode \\
\midrule
Qwen2.5-3B-Instruct   & 3B & 36 & 2048 & standard \\
Qwen3-4B              & 4B & 36 & 2560 & standard \\
Qwen3-4B              & 4B & 36 & 2560 & thinking \\
Llama-3.2-3B-Instruct & 3B & 28 & 3072 & standard \\
Llama-3.1-8B-Instruct & 8B & 32 & 4096 & standard \\
\bottomrule
\end{tabular}
\end{center}
\caption{Models evaluated. All models are loaded in bf16 full precision with
frozen weights. The thinking-mode variant of Qwen3-4B differs from the standard
variant solely in having extended chain-of-thought reasoning enabled at
inference time.}
\label{tab:models}
\end{table}
\subsection{Contrastive probing design}
\label{sec:design}

The central choice is a contrastive pairing within each base problem $i$: a clean
context $\mathbf{x}_i^{(0)}$ and a corrupted context $\mathbf{x}_i^{(1)}$ differing
only in the embedded hop-0 answer (one replaced by a synthetically wrong value),
sharing question, hop structure, and hop-1 prompt. Representation differences are
thus attributable to the injected error, controlling for difficulty and hop
structure and ruling out a probe that scores high by learning difficulty rather
than error presence.\footnote{This controls for problem identity, not every
surface property: the contexts still differ in the literal corrupted value, so
some surface cues (digits, units, magnitude, length) correlate with the label by
construction. We test generalisation beyond surface artifacts with a
leave-one-error-type-out evaluation (\Cref{app:heldout_type}).} Each base problem
contributes three conditions: \textbf{Clean} ($y=0$); \textbf{Error-standard}
($y=1$), error at hop 0, the primary probing condition; and
\textbf{Error-verbalized} ($y=1$), identical but eliciting a confidence score, to
test whether elicitation changes behavior independently of internal state. Two
further passes yield \texttt{final\_degraded} -- whether the injection produced a
wrong final answer on an otherwise-solvable problem -- the target for failure
prediction (\Cref{sec:stage3}) and intervention evaluation (\Cref{sec:stage4}).

\subsection{Probe robustness controls}
\label{sec:robustness}

Two controls test whether the probe captures an operational error state rather
than memorised surface structure. \textbf{Grouped cross-validation} replaces the
stratified folds of \Cref{sec:stage2} with five folds that hold out entire base
problems, so no template or near-duplicate context is shared across train and
test. \textbf{Held-out error type} trains the probe on four of the five error
types and evaluates detection on the fifth, repeated for all five leave-one-out
splits; above-chance transfer to a never-seen surface corruption indicates a
corruption feature that generalises across surface forms rather than a per-type
token signature. Both controls reuse the best aggregation mode from
\Cref{sec:stage2}; the grouped analysis selects its layer within the grouped
folds. Full results are in \Cref{app:grouped_cv,app:heldout_type}.

\subsection{Activation extraction}
\label{sec:activations}

Each forward pass captures the full stack of hidden states
$\{\mathbf{h}^{(\ell)}\}_{\ell=0}^{L}$ at every layer $\ell$, including the
embedding layer at $\ell = 0$. For a prompt of $T$ tokens, each
$\mathbf{h}^{(\ell)} \in \mathbb{R}^{T \times d_\text{model}}$. Two aggregations
produce fixed-size representation vectors per layer:

\begin{equation}
\begin{aligned}
  \mathbf{a}^{(\ell)}_\text{last} &= \mathbf{h}^{(\ell)}_{T}
  \in \mathbb{R}^{d_\text{model}},
  \\
  \mathbf{a}^{(\ell)}_\text{mean} &= \frac{1}{T}
  \sum_{t=1}^{T} \mathbf{h}^{(\ell)}_{t}
  \in \mathbb{R}^{d_\text{model}}.
\end{aligned}
  \label{eq:agg}
\end{equation}

The last-token aggregation captures the representation at the position that
conditions autoregressive generation, and is the standard choice in the probing
literature \citep{belinkov2022probing,gurnee2023finding}. The mean-over-tokens
aggregation captures a whole-prompt summary signal. We include both because
deeper models may distribute the error signal across token positions rather than
concentrating it at the final token; selecting the better aggregation per layer
is part of the model selection procedure described in \Cref{sec:stage2}.

We visualise how the mean-pooled signal accumulates across token positions in
\Cref{app:trajectory}, which clarifies why mean pooling is informative even at
shallow layers.

\subsection{Surface uncertainty signals}
\label{sec:surface}

In parallel with activation extraction, six scalar uncertainty signals are
computed from the per-token logit distribution $p_t$. Let $p_t^{(1)} \geq
p_t^{(2)} \geq \ldots$ be the sorted token probabilities at position $t$, let
$H(p_t) = -\sum_v p_t(v) \log p_t(v)$ be the Shannon entropy, and $G$ the number
of generated tokens:
\begin{equation}
\begin{aligned}
  s_\text{peak-ent}   &= \max_{t}\, H(p_t), \\
  s_\text{mean-ent}   &= \frac{1}{G}\sum_{t=1}^{G} H(p_t), \\
  s_\text{early-ent}  &= \frac{1}{\min(G,20)}
                         \sum_{t=1}^{\min(G,20)} H(p_t), \\
  s_\text{min-conf}   &= \min_{t}\, p_t^{(1)}, \\
  s_\text{vocab-gap}  &= \frac{1}{G}\sum_{t=1}^{G}
                         \bigl(p_t^{(1)} - p_t^{(2)}\bigr), \\
  s_\text{logprob}    &= -\frac{1}{G}\sum_{t=1}^{G}
                         \log p_t(y_t).
\end{aligned}
  \label{eq:signals}
\end{equation}
These span three hypotheses about where uncertainty surfaces: distributional
spread (entropy), top-token certainty (confidence), and top-two margin
(vocabulary gap). Together they form the surface baselines against which the probe
is compared.\footnote{For the AUROC comparisons (\Cref{eq:auroc,eq:detect}),
$s_\text{min-conf}$ and $s_\text{vocab-gap}$ are sign-flipped so larger values
indicate the error class; reported sub-0.5 values are therefore genuinely
non-predictive rather than orientation artifacts. Signal interpretations and the
four behavioral text flags are detailed in \Cref{app:surface_signals}.}

\subsection{Layer-wise linear probe}
\label{sec:stage2}

A linear probe is trained independently at every layer $\ell \in \{0, \ldots,
L\}$ under both aggregation modes. The probe is a logistic regression classifier
with input standardisation:

\begin{equation}
  \hat{y} = \sigma\!\left(\mathbf{w}^\top
    \frac{\mathbf{a} - \boldsymbol{\mu}}{\boldsymbol{\sigma}} + b\right),
  \label{eq:probe}
\end{equation}

where $\boldsymbol{\mu}$ and $\boldsymbol{\sigma}$ are the per-feature mean and
standard deviation estimated from the training split, $\mathbf{w} \in
\mathbb{R}^{d_\text{model}}$ and $b \in \mathbb{R}$ are the learned weights and
bias, and $\sigma(\cdot)$ is the sigmoid function.

We use a linear probe deliberately. A nonlinear classifier could achieve high
accuracy by learning complex decision boundaries that do not correspond to any
interpretable direction in the residual stream. A linear probe succeeds only if
the error signal is \emph{linearly decodable}, that is, if clean and corrupted
representations are linearly separable in activation space. This is the standard
operationalisation of the claim that the model has encoded a feature in the
mechanistic interpretability literature
\citep{belinkov2022probing,gurnee2023finding}.

\paragraph{Layer selection.}
Each layer/mode combination is evaluated under 5-fold stratified
cross-validation, scored by AUROC:
\begin{equation}
  \text{AUROC} = \Pr\!\left[\hat{y}(\mathbf{a}^+) > \hat{y}(\mathbf{a}^-)\right],
  \label{eq:auroc}
\end{equation}
for a randomly drawn error sample $\mathbf{a}^+$ and clean sample $\mathbf{a}^-$,
reported as mean$\pm$std across folds. The highest-mean-AUROC layer/mode is the
best layer; a final probe is fit on all $N$ samples there for downstream
stages.\footnote{Shared base problems and templates mean stratified folds may
leak near-duplicate contexts across train/test; the contrastive control
(\Cref{sec:design}) mitigates but does not eliminate this, so we additionally run
grouped cross-validation holding out entire base problems (\Cref{app:grouped_cv});
detection survives this stricter control.}

\paragraph{Threshold calibration.}
The decision threshold $\tau$ is calibrated on out-of-fold scores. Three modes are
derived. By construction $\tau_\text{loose} = 0.7\,\tau_\text{Youden} <
\tau_\text{Youden}$; the relation of $\tau_\text{conservative}$ to
$\tau_\text{Youden}$ is empirical, and on the highly separable score distributions
we observe the two nearly coincide (\Cref{app:thresholds}):
\begin{equation}
\begin{aligned}
  \tau_\text{Youden} &= \arg\max_\tau
    \bigl[\text{TPR}(\tau) - \text{FPR}(\tau)\bigr],
  \\
  \tau_\text{conservative} &= F^{-1}_{y=0}(0.99),
  \\
  \tau_\text{loose} &= 0.7\,\tau_\text{Youden}.
\end{aligned}
  \label{eq:thresholds}
\end{equation}
By \Cref{eq:fire}, a lower threshold fires more aggressively.
$\tau_\text{Youden}$ is the default in all intervention
experiments; the sensitivity ablation (\Cref{sec:sensitivity}) sweeps all three.

\subsection{Detection versus failure prediction}
\label{sec:stage3}

We compute two AUROC scores per signal $s$:
\begin{align}
  \text{AUROC}_\text{detect}(s) &= \Pr\bigl[s(\mathbf{x}^{(1)}) >
    s(\mathbf{x}^{(0)})\bigr],
    \label{eq:detect} \\
  \text{AUROC}_\text{fail}(s)   &= \Pr\bigl[s(\mathbf{x}^{(1)}_\text{degraded})
    \nonumber \\ &\qquad
    > s(\mathbf{x}^{(1)}_\text{recovered})\bigr].
    \label{eq:fail}
\end{align}
\Cref{eq:detect} tests \emph{detection} across all records; \Cref{eq:fail}, over
the error subset only, tests \emph{failure prediction}, using the
\texttt{final\_degraded} target (wrong under error, correct under matched clean)
to isolate propagation from baseline difficulty. The two need not correlate: a
signal can detect perfectly yet be uninformative about failure if the error
representation saturates regardless of downstream recovery. This is the
methodological core of the paper. Prior calibration work asks whether verbalized
confidence correlates with accuracy; we ask the prior question, whether the
\emph{internal} detection signal also predicts propagation, which determines
whether probe-based monitoring is actionable.

For verbalized confidence, binary collapse is:
\begin{equation}
\begin{aligned}
  \text{collapse rate} = \frac{1}{N_\text{verb}}
    \sum_{i=1}^{N_\text{verb}}
    \mathbf{1}\bigl[&c_i \leq 0.05 \\[-2pt]
    &\;\text{or}\; c_i \geq 0.95\bigr],
\end{aligned}
  \label{eq:collapse}
\end{equation}
with $c_i = v_i/10$; wrong rates of the two extreme groups are then compared to
test whether the split carries any correctness signal.

\subsection{Intervention experiment}
\label{sec:stage4}

The intervention experiment tests whether the probe can serve as a runtime monitor
that improves final-answer accuracy. The probe fires on trace $i$ at hop $k$ when
its output exceeds the Youden threshold:
\begin{equation}
  \text{fire}(i, k) = \mathbf{1}\!\left[
    \sigma\!\left(\mathbf{w}^\top
    \frac{\mathbf{a}_k^{(i)} - \boldsymbol{\mu}}{\boldsymbol{\sigma}}
    + b\right) \geq \tau_\text{Youden}
  \right].
  \label{eq:fire}
\end{equation}
On firing, one of three strategies is applied, subject to a cap of $M = 3$ per
trace to bound inference cost. \textbf{Reprompt (C1)} augments the context with a
fixed recheck instruction before generation; it is the cheapest policy, adding no
forward passes, and tests whether a verbal alert alone changes behaviour.
\textbf{Replace-prior (C2)} regenerates hop $k-1$ from scratch,
$\tilde{r}_{k-1} = f_\theta(q, r_0, \ldots, r_{k-2})$, substitutes it for the
corrupted value in the context, and continues generation from the corrected
prefix $(q, r_0, \ldots, r_{k-2}, \tilde{r}_{k-1})$, directly removing the
corruption at the cost of one extra forward pass per firing.
\textbf{Branch-and-pick (C3)} samples $K$ candidate hop-$k$ responses
$\tilde{r}_k^{(1)}, \ldots, \tilde{r}_k^{(K)}$ at temperatures
$\mathcal{T} = \{0.7, 0.85, 1.0, 1.15\}$ from the committed prefix
$r^{*}_{0:k-1}$, and selects the one inducing the least error-like internal state,
$j^{*} = \arg\min_{j \in [K]} \hat{y}\bigl(\mathbf{a}(q, r^{*}_{0:k-1},
\tilde{r}_k^{(j)})\bigr)$, using the probe to pick the continuation least likely to
propagate the error, at a cost of $K$ extra forward passes per firing.

\subsection{Sensitivity ablation}
\label{sec:sensitivity}

We evaluate branch-and-pick across the full factorial of branch count and
threshold mode, $K \in \{1, 2, 4, 8\}$ crossed with $\tau \in
\{\tau_\text{conservative}, \tau_\text{Youden}, \tau_\text{loose}\}$ (12 cells per
model), exposing the precision/recall tradeoff in the branch decision: looser
thresholds and larger $K$ rescue more errors at the cost of higher broken counts.
Sampling details are given in \Cref{app:sensitivity}.

\section{Results}
\label{sec:results}

\paragraph{Probe detects corruption but does not predict failure.}
\label{sec:probe_results}
Across all five variants, a linear probe on residual-stream activations exceeds
$\auroc_\text{detect} > 0.98$ at the best layer (\Cref{tab:probe_summary}),
peaking at $0.997$ on Llama-3.1-8B under mean pooling. All models report
$\auroc = 0.500$ at $\ell=0$ under last-token pooling, so the signal arises from
computation rather than token identity; a single layer-0$\to$1 jump of
$0.38$--$0.41$ AUROC contributes the bulk of it. Peak depth grows with model size
(39\% of depth for Llama-3.2-3B, 64\% for Qwen3-4B instruct). Crucially, the same
probe is \emph{uninformative about failure}: at its best detection layer its
failure-prediction AUROC (\Cref{eq:fail}) collapses to near chance across all five
models, in sharp contrast to detection above $0.98$. This gap between
$\auroc_\text{detect}$ and $\auroc_\text{fail}$ within a single probe is the core
dissociation of the paper; per-model values are in \Cref{tab:probe_summary}.

\begin{table*}[t]
\begin{center}
\small
\begin{tabular}{lccccc}
\toprule
Model & Best layer (last) & AUROC (last) & AUROC (mean) & $\auroc_\text{fail}$ & Best surface \\
\midrule
Llama-3.2-3B      & 11/28 & 0.985 & 0.980 & 0.49 & 0.574 \\
Llama-3.1-8B      & 15/32 & 0.993 & \textbf{0.997} & 0.53 & 0.622 \\
Qwen2.5-3B        & 22/36 & \textbf{0.994} & 0.990 & 0.51 & 0.587 \\
Qwen3-4B (inst.)  & 23/36 & 0.990 & 0.993 & 0.47 & 0.637 \\
Qwen3-4B (think)  & 19/36 & 0.989 & 0.989 & 0.50 & 0.694 \\
\bottomrule
\end{tabular}
\end{center}
\caption{Best-layer probe detection AUROC (last/mean pooling), probe
\emph{failure-prediction} AUROC at the same layer, and the best surface signal.
Detection exceeds $0.98$ everywhere while $\auroc_\text{fail}$ is at or near
chance, and the detection gap to the best surface signal is at least $0.30$.}
\label{tab:probe_summary}
\end{table*}

\paragraph{Surface signals are at or near chance.}
\label{sec:surface_results}
None of the surface signals shown exceeds $\auroc = 0.70$ for detection
(\Cref{tab:surface}, \Cref{app:surface_thinking}); peak entropy is exactly
$0.500$ for the instruct models shown, and the mean probe-to-surface gap is
$0.41$. The
Qwen3-4B thinking model reverses below chance ($0.308$), with error traces
\emph{less} entropic than clean ones, consistent with thinking-mode suppression
of surface uncertainty (\Cref{sec:thinking_ablation}).

\paragraph{Behavioral silence.}
\label{sec:behavioral}
Despite near-perfect internal detection, all four instruct models produce zero
hedging and zero overconfident outputs across 258 traces each, and never use a
structured confidence format unprompted; thinking models hedge only 1--2\% of the
time. Clean-condition accuracy is 23--31\% and degrades 4--8 points under
injection. The \texttt{error\_verbalized} condition does not improve accuracy
over \texttt{error\_standard} in any model.

\paragraph{Thinking mode degrades accuracy without changing encoding.}
\label{sec:thinking_ablation}
Enabling chain-of-thought on Qwen3-4B barely moves probe AUROC
($\Delta = -0.001$ last, $-0.004$ mean) but degrades error-condition accuracy
from 27.9\% to 1.2\% (\Cref{tab:thinking}, \Cref{app:surface_thinking}), despite
$\sim$5$\times$ more tokens. The variants share weights, differing only in the
inference-time template, so the encoding persists while the output channel is
disrupted. Clean accuracy also drops (29.1\% to 12.8\%), so this best evidences
\emph{persistence} of the encoding under degradation rather than an isolated
manipulation of the saying channel.

\paragraph{Interventions: branch-and-pick is closest to asymmetric.}
\label{sec:intervention_results}
Across three models with complete data, branch-and-pick is net-positive everywhere
and uniquely non-breaking on Llama-3.1-8B (4 rescued, 0 broken), breaking one trace
on Qwen2.5-3B; replace-prior breaks correct traces at nearly the rate it rescues
wrong ones, and reprompt does neither. The probe fires on 96--100\% of error traces,
yet replace-prior's firing drops to 32--39\% with no correctness gain, so the
encoding is not the proximal cause of wrong answers. Per-type effects are strongly
heterogeneous -- replace-prior best on \texttt{wrong\_unit} ($+37.5\%$),
net-negative on \texttt{wrong\_percentage\_base} ($-18.2\%$) -- motivating
error-type-aware routing, exploratory given 9--16 examples per slice
(\Cref{tab:byerror}, \Cref{app:per_error}).

\section{Evaluation metrics}
\label{sec:metrics}

We evaluate three tasks, each with its own metric.

\paragraph{Detection and failure prediction.}
For the probe and all surface signals we report AUROC (\Cref{eq:auroc}), which is
threshold-free, with $0.5$ denoting chance and values below $0.5$ indicating an
anti-predictive (inverted) signal. It is
computed for two targets:
\begin{equation}
\begin{aligned}
  y_\text{detect} &= \mathbf{1}[\text{error injected}],
  \\
  y^{(i)}_\text{fail} &= \mathbf{1}\bigl[\hat{a}^{(i)}_\text{err} \neq a^{*}_i
    \;\wedge\; \hat{a}^{(i)}_\text{clean} = a^{*}_i\bigr],
\end{aligned}
\end{equation}
over all traces (\Cref{eq:detect}) and the error subset (\Cref{eq:fail})
respectively, where $\hat{a}^{(i)}_\text{err}$ and $\hat{a}^{(i)}_\text{clean}$ are
the model's final answers on trace $i$ under the error and matched-clean contexts
and $a^{*}_i$ is ground truth. This \texttt{final\_degraded} target separates
error propagation from baseline difficulty. The two are decoupled by design:
$\text{AUROC}_\text{detect}\!\approx\!1$ with
$\text{AUROC}_\text{fail}\!\approx\!0.5$ is the central measurement target.

\paragraph{Verbalization quality.}
With $c_i = v_i/10$ the normalised confidence and collapse rate as in
\Cref{eq:collapse}, we compare the wrong rate of the two collapsed groups
$g \in \{[0,0.05],[0.95,1]\}$:
\begin{equation}
\begin{aligned}
  \text{wrong}(g) &= \frac{1}{|G_g|}\sum_{i \in G_g}
    \mathbf{1}[\hat{a}_i \neq a_i^*],
  \\
  G_g &= \{i : c_i \in g\}.
\end{aligned}
\end{equation}
Indistinguishable $\text{wrong}([0,0.05])$ and $\text{wrong}([0.95,1])$ confirm
the binary collapse carries no calibration signal.

\paragraph{Intervention effectiveness.}
For each policy $c \in \{\text{C1},\text{C2},\text{C3}\}$, relative to baseline:
\begin{equation}
\begin{aligned}
  \text{rescued}_c &= |\{i : \text{wrong}^{(i)}_\text{base}
    \wedge \text{correct}^{(i)}_c\}|, \\
  \text{broken}_c &= |\{i : \text{correct}^{(i)}_\text{base}
    \wedge \text{wrong}^{(i)}_c\}|, \\
  \text{net}_c &= \text{rescued}_c - \text{broken}_c, \\
  \Delta\text{acc}_c &= \tfrac{\text{net}_c}{N_\text{intervene}} \times 100.
\end{aligned}
\end{equation}
Rescued and broken are reported separately, not collapsed into net, since equal
net can hide very different risk profiles; an asymmetric policy
($\text{rescued} \gg \text{broken}$) is the deployment target. Answers match
ground truth within $10^{-3}$ absolute or $1\%$ relative tolerance.

\section{Discussion}
\label{sec:discussion}

\paragraph{Robustness to leakage and surface artifacts.}
Detection is not an artifact of fold leakage or per-type surface cues. Under
grouped cross-validation holding out entire base problems, best-layer detection
AUROC moves by at most $0.003$ from the stratified estimate (\Cref{app:grouped_cv}).
Under leave-one-error-type-out training, detection on the unseen type stays above
$0.91$ in every model-by-type cell (per-model means $0.968$--$0.996$;
\Cref{app:heldout_type}). \texttt{off\_by\_one}, least separable on surface
features yet still well above chance, is the strongest evidence that the probe
encodes an abstract corruption state rather than a per-type token signature.

\paragraph{Decodability, not causation.}
We claim decodability, not causation: the corruption signal is linearly present in
the residual stream, but this does not establish that the generation policy
consults that direction. Our interventions (\Cref{sec:intervention_results}) bear
on this -- replace-prior removes the corrupted value and sharply lowers probe
firing (96--100\% to 32--39\%) without a proportional accuracy gain, evidence that
the probed direction is not necessarily the proximal cause of the wrong answer.
Establishing causal relevance would require representation-level interventions
(activation patching, mediation analysis), which we leave to future work.

\paragraph{Toward routed interventions.}
The per-error-type results (\Cref{sec:intervention_results},
\Cref{app:per_error}) show that no single intervention dominates. Replace-prior is
strongest on \texttt{wrong\_unit} but net-negative on
\texttt{wrong\_percentage\_base}, while branch-and-pick is the only strictly
net-positive policy on \texttt{off\_by\_one}. This motivates a routed policy that
selects the intervention from a predicted error type. We stress that such routing
presupposes an error-type classifier, which we do not yet have, and that the
supporting per-type sample sizes are small, so we frame routing as a direction
rather than a deployable result.

\paragraph{Future work.}
Three directions look most promising. Per-hop localisation would move the probe
target from ``a prior is corrupted'' to ``hop $k$ is corrupted,'' enabling much
more targeted replace-prior. Error-type probes would classify the corruption
itself, unlocking the routed policy without requiring an oracle. And mechanistic
dissection of the layer 0 to layer 1 jump would clarify whether the signal arises
from attention to inconsistent tokens or from an MLP feature. To our knowledge
this is an unusually clean single-layer information gain and worth understanding.

\section{Conclusion}
We characterise a four-part dissociation in five models reasoning under silently
corrupted context. A linear probe detects injected errors near-perfectly
($\auroc > 0.98$, up to $0.997$ on Llama-3.1-8B) yet does not predict whether the
error reaches the final answer; verbalised confidence collapses to a binary filter
with indistinguishable wrong rates; probe persistence across hops does not separate
outcomes, refuting our pre-registered persistence-beats-peak hypothesis; and
chain-of-thought widens rather than closes the gap, leaving probe \auroc{} unchanged
while error-condition accuracy falls to 1.2\%. No intervention dominates:
branch-and-pick is net-positive everywhere and uniquely non-breaking on
Llama-3.1-8B, but the asymmetry is conditional on error type and model
(\Cref{sec:intervention_results,app:per_error}). Probe-based monitoring is thus a
necessary complement to verbalised confidence rather than a replacement, and the
deployable answer is model-aware, error-type-aware routing.

\section*{Ethics Statement}

This paper presents a probe-based mechanism for detecting and intervening on
errors during multi-step reasoning by language models. The intended use is
trustworthy deployment, namely catching errors before commitment to a wrong
answer. Three considerations apply. First, the probe operates on internal
activations and cannot be used by parties without access to them, so it is not a
black-box method. Second, the interventions are not free of failure modes.
Replace-prior breaks previously-correct traces at nearly the rate it rescues
wrong ones, and branch-and-pick breaks at a small but non-zero rate, so
probe-triggered intervention should not be presented to end users as strictly
safe. Third, branch-and-pick increases inference cost roughly tenfold at the
intervention frequencies measured here, with energy and latency implications
worth weighing for any specific application.

\bibliography{references}

\begin{thebibliography}{32}
\providecommand{\natexlab}[1]{#1}

\bibitem[{{Anthropic}(2024{\natexlab{a}})}]{anthropic2024claude}
{Anthropic}. 2024{\natexlab{a}}.
\newblock The {Claude} 3 model family: {Opus}, {Sonnet}, {Haiku}.
\newblock
  \url{https://www-cdn.anthropic.com/de8ba9b01c9ab7cbabf5c33b80b7bbc618857627/Model_Card_Claude_3.pdf}.

\bibitem[{{Anthropic}(2024{\natexlab{b}})}]{anthropic2024claudecode}
{Anthropic}. 2024{\natexlab{b}}.
\newblock {Claude Code}: Agentic coding.
\newblock \url{https://www.anthropic.com/product/claude-code}.

\bibitem[{Arditi et~al.(2024)Arditi, Obeso, Syed, Paleka, Panickssery, Gurnee,
  and Nanda}]{arditi2024refusal}
Andy Arditi, Oscar Obeso, Aaquib Syed, Daniel Paleka, Nina Panickssery, Wes
  Gurnee, and Neel Nanda. 2024.
\newblock Refusal in language models is mediated by a single direction.
\newblock In \emph{Advances in Neural Information Processing Systems 37
  (NeurIPS 2024)}.

\bibitem[{Belinkov(2022)}]{belinkov2022probing}
Yonatan Belinkov. 2022.
\newblock Probing classifiers: Promises, shortcomings, and advances.
\newblock \emph{Computational Linguistics}, 48(1):207--219.

\bibitem[{Burns et~al.(2023)Burns, Ye, Klein, and Steinhardt}]{burns2023latent}
Collin Burns, Haotian Ye, Dan Klein, and Jacob Steinhardt. 2023.
\newblock Discovering latent knowledge in language models without supervision.
\newblock In \emph{The Eleventh International Conference on Learning
  Representations (ICLR)}.

\bibitem[{Chan et~al.(2024)Chan, Chen, Su, Yu, Xue, Zhang, Fu, and
  Liu}]{chan2023chateval}
Chi-Min Chan, Weize Chen, Yusheng Su, Jianxin Yu, Wei Xue, Shanghang Zhang, Jie
  Fu, and Zhiyuan Liu. 2024.
\newblock {ChatEval}: Towards better {LLM}-based evaluators through multi-agent
  debate.
\newblock In \emph{International Conference on Learning Representations}.

\bibitem[{{Cognition AI}(2024)}]{cognition2024devin}
{Cognition AI}. 2024.
\newblock Introducing devin, the first {AI} software engineer.
\newblock \url{https://cognition.ai/blog/introducing-devin}.

\bibitem[{Gurnee et~al.(2023)Gurnee, Nanda, Pauly, Harvey, Troiano, and
  Steinhardt}]{gurnee2023finding}
Wes Gurnee, Neel Nanda, Matthew Pauly, Kyle Harvey, Dmitrii Troiano, and Jacob
  Steinhardt. 2023.
\newblock Finding neurons in a haystack: Case studies with sparse probing.
\newblock \emph{Transactions on Machine Learning Research}.

\bibitem[{Hewitt and Manning(2019)}]{hewitt2019structural}
John Hewitt and Christopher~D. Manning. 2019.
\newblock A structural probe for finding syntax in word representations.
\newblock In \emph{Proceedings of the 2019 Conference of the North American
  Chapter of the Association for Computational Linguistics (NAACL)}, pages
  4129--4138.

\bibitem[{Huang et~al.(2023)Huang, Yu, Ma, Zhong, Feng, Wang, Chen, Peng, Feng,
  Qin et~al.}]{huang2023survey}
Lei Huang, Weijiang Yu, Weitao Ma, Weihong Zhong, Zhengwei Feng, Haotian Wang,
  Qianglong Chen, Weihua Peng, Xiaocheng Feng, Bing Qin, et~al. 2023.
\newblock A survey on hallucination in large language models: Principles,
  taxonomy, challenges, and open questions.
\newblock \emph{arXiv preprint arXiv:2311.05232}.

\bibitem[{Jimenez et~al.(2024)Jimenez, Yang, Wettig, Yao, Pei, Press, and
  Narasimhan}]{jimenez2024swebench}
Carlos~E Jimenez, John Yang, Alexander Wettig, Shunyu Yao, Kexin Pei, Ofir
  Press, and Karthik Narasimhan. 2024.
\newblock {SWE-bench}: Can language models resolve real-world {GitHub} issues?
\newblock In \emph{International Conference on Learning Representations}.

\bibitem[{Kadavath et~al.(2022)Kadavath, Conerly, Askell, Henighan, Ganguli,
  Mirhoseini, Jones, Chen, LightmanHarrison et~al.}]{kadavath2022know}
Saurav Kadavath, Tom Conerly, Amanda Askell, Tom Henighan, Deep Ganguli, Azalia
  Mirhoseini, Andy Jones, Anna Chen, LightmanHarrison, et~al. 2022.
\newblock Language models (mostly) know what they know.
\newblock \emph{arXiv preprint arXiv:2207.05221}.

\bibitem[{Lanham et~al.(2023)Lanham, Chen, Radhakrishnan, Steiner, Denison,
  Hernandez, Li, Durmus, Hubinger, Kernion et~al.}]{lanham2023measuring}
Tamera Lanham, Anna Chen, Ansh Radhakrishnan, Benoit Steiner, Carson Denison,
  Danny Hernandez, Dustin Li, Esin Durmus, Evan Hubinger, Jackson Kernion,
  et~al. 2023.
\newblock Measuring faithfulness in chain-of-thought reasoning.
\newblock \emph{arXiv preprint arXiv:2307.13702}.

\bibitem[{Lightman et~al.(2024)Lightman, Kosaraju, Burda, Edwards, Baker, Lee,
  Leike, Schulman, Sutskever, and Cobbe}]{lightman2024lets}
Hunter Lightman, Vineet Kosaraju, Yura Burda, Harri Edwards, Bowen Baker, Teddy
  Lee, Jan Leike, John Schulman, Ilya Sutskever, and Karl Cobbe. 2024.
\newblock Let's verify step by step.
\newblock In \emph{The Twelfth International Conference on Learning
  Representations (ICLR)}.

\bibitem[{Lin et~al.(2022)Lin, Hilton, and Evans}]{lin2022teaching}
Stephanie Lin, Jacob Hilton, and Owain Evans. 2022.
\newblock Teaching models to express their uncertainty in words.
\newblock \emph{Transactions on Machine Learning Research}.

\bibitem[{Marks and Tegmark(2023)}]{marks2023geometry}
Samuel Marks and Max Tegmark. 2023.
\newblock The geometry of truth: Emergent linear structure in large language
  model representations of true/false datasets.
\newblock \emph{arXiv preprint arXiv:2310.06824}.

\bibitem[{{OpenAI}(2023)}]{openai2024gpt4}
{OpenAI}. 2023.
\newblock {GPT-4} technical report.
\newblock \url{https://arxiv.org/abs/2303.08774}.

\bibitem[{Pan et~al.(2023)Pan, Saxon, Xu, Nathani, Wang, and
  Wang}]{pan2023automatically}
Liangming Pan, Michael Saxon, Wenda Xu, Deepak Nathani, Xinyi Wang, and
  William~Yang Wang. 2023.
\newblock Automatically correcting large language models: Surveying the
  landscape of diverse automated correction strategies.
\newblock \emph{Transactions of the Association for Computational Linguistics},
  11:1409--1429.

\bibitem[{Paul et~al.(2024)Paul, West, Bosselut, and Faltings}]{paul2024making}
Debjit Paul, Robert West, Antoine Bosselut, and Boi Faltings. 2024.
\newblock Making reasoning matter: Measuring and improving faithfulness of
  chain-of-thought reasoning.
\newblock In \emph{Findings of the Association for Computational Linguistics:
  EMNLP 2024}.

\bibitem[{Schick et~al.(2023)Schick, Dwivedi-Yu, Dess{\`i}, Raileanu, Lomeli,
  Zettlemoyer, Cancedda, and Scialom}]{schick2023toolformer}
Timo Schick, Jane Dwivedi-Yu, Roberto Dess{\`i}, Roberta Raileanu, Maria
  Lomeli, Luke Zettlemoyer, Nicola Cancedda, and Thomas Scialom. 2023.
\newblock Toolformer: Language models can teach themselves to use tools.
\newblock In \emph{Advances in Neural Information Processing Systems},
  volume~36.

\bibitem[{Shinn et~al.(2023)Shinn, Cassano, Gopinath, Narasimhan, and
  Yao}]{shinn2023reflexion}
Noah Shinn, Federico Cassano, Ashwin Gopinath, Karthik Narasimhan, and Shunyu
  Yao. 2023.
\newblock Reflexion: Language agents with verbal reinforcement learning.
\newblock In \emph{Advances in Neural Information Processing Systems},
  volume~36.

\bibitem[{Singhal et~al.(2023)Singhal, Azizi, Tu, Mahdavi, Wei, Chung, Scales,
  Tanwani, Cole-Lewis, Pfohl et~al.}]{singhal2023large}
Karan Singhal, Shekoofeh Azizi, Tao Tu, S~Sara Mahdavi, Jason Wei, Hyung~Won
  Chung, Nathan Scales, Ajay Tanwani, Heather Cole-Lewis, Stephen Pfohl, et~al.
  2023.
\newblock Large language models encode clinical knowledge.
\newblock In \emph{Nature}, volume 620, pages 172--180.

\bibitem[{Tian et~al.(2023)Tian, Mitchell, Zhou, Sharma, Rafailov, Yao, Finn,
  and Manning}]{tian2023ask}
Katherine Tian, Eric Mitchell, Allan Zhou, Archit Sharma, Rafael Rafailov,
  Huaxiu Yao, Chelsea Finn, and Christopher~D. Manning. 2023.
\newblock Just ask for calibration: Strategies for eliciting calibrated
  confidence scores from language models fine-tuned with human feedback.
\newblock In \emph{Proceedings of the 2023 Conference on Empirical Methods in
  Natural Language Processing (EMNLP)}.

\bibitem[{Turpin et~al.(2023)Turpin, Michael, Perez, and
  Bowman}]{turpin2023language}
Miles Turpin, Julian Michael, Ethan Perez, and Samuel~R. Bowman. 2023.
\newblock Language models don't always say what they think: Unfaithful
  explanations in chain-of-thought prompting.
\newblock In \emph{Advances in Neural Information Processing Systems 36
  (NeurIPS 2023)}.

\bibitem[{Valmeekam et~al.(2023)Valmeekam, Marquez, Sreedharan, and
  Kambhampati}]{valmeekam2023planning}
Karthik Valmeekam, Matthew Marquez, Sarath Sreedharan, and Subbarao
  Kambhampati. 2023.
\newblock Large language models still can't plan.
\newblock In \emph{NeurIPS 2023 Foundation Models for Decision Making
  Workshop}.

\bibitem[{Wang et~al.(2024)Wang, Ma, Feng, Zhang, Yang, Zhang, Chen, Tang,
  Chen, Lin et~al.}]{wang2024survey}
Lei Wang, Chen Ma, Xueyang Feng, Zeyu Zhang, Hao Yang, Jingsen Zhang, Zhiyuan
  Chen, Jiakai Tang, Xu~Chen, Yankai Lin, et~al. 2024.
\newblock A survey on large language model based autonomous agents.
\newblock \emph{Frontiers of Computer Science}, 18(6):186345.

\bibitem[{Wang et~al.(2023)Wang, Wei, Schuurmans, Le, Chi, Narang, Chowdhery,
  and Zhou}]{wang2023self}
Xuezhi Wang, Jason Wei, Dale Schuurmans, Quoc~V. Le, Ed~H. Chi, Sharan Narang,
  Aakanksha Chowdhery, and Denny Zhou. 2023.
\newblock Self-consistency improves chain of thought reasoning in language
  models.
\newblock In \emph{The Eleventh International Conference on Learning
  Representations (ICLR)}.

\bibitem[{Wei et~al.(2022)Wei, Wang, Schuurmans, Bosma, Ichter, Xia, Chi, Le,
  and Zhou}]{wei2022chain}
Jason Wei, Xuezhi Wang, Dale Schuurmans, Maarten Bosma, Brian Ichter, Fei Xia,
  Ed~Chi, Quoc Le, and Denny Zhou. 2022.
\newblock Chain-of-thought prompting elicits reasoning in large language
  models.
\newblock In \emph{Advances in Neural Information Processing Systems},
  volume~35.

\bibitem[{Weng et~al.(2023)Weng, Zhu, Xia, Li, He, Liu, Sun, Liu, and
  Zhao}]{weng2023large}
Yixuan Weng, Minjun Zhu, Fangyu Xia, Bin Li, Shizhu He, Shengping Liu, Bin Sun,
  Kang Liu, and Jun Zhao. 2023.
\newblock Large language models are better reasoners with self-verification.
\newblock \emph{arXiv preprint arXiv:2212.09561}.

\bibitem[{Xiong et~al.(2024)Xiong, Hu, Lu, Li, Fu, He, and Hooi}]{xiong2024can}
Miao Xiong, Zhiyuan Hu, Xinyang Lu, Yifei Li, Jie Fu, Junxian He, and Bryan
  Hooi. 2024.
\newblock Can {LLMs} express their uncertainty? an empirical evaluation of
  confidence elicitation in {LLMs}.
\newblock \emph{arXiv preprint arXiv:2306.13063}.

\bibitem[{Yang et~al.(2024)Yang, Jimenez, Wettig, Lieret, Yao, Narasimhan, and
  Press}]{yang2024sweagent}
John Yang, Carlos~E Jimenez, Alexander Wettig, Kilian Lieret, Shunyu Yao,
  Karthik Narasimhan, and Ofir Press. 2024.
\newblock {SWE-agent}: Agent-computer interfaces enable automated software
  engineering.
\newblock In \emph{Advances in Neural Information Processing Systems},
  volume~37.

\bibitem[{Yao et~al.(2023)Yao, Zhao, Yu, Du, Shafran, Narasimhan, and
  Cao}]{yao2023react}
Shunyu Yao, Jeffrey Zhao, Dian Yu, Nan Du, Izhak Shafran, Karthik Narasimhan,
  and Yuan Cao. 2023.
\newblock {ReAct}: Synergizing reasoning and acting in language models.
\newblock In \emph{International Conference on Learning Representations}.

\end{thebibliography}

\newpage
\appendix
\section*{Appendix: Supplementary Methodology and Results}
\addcontentsline{toc}{section}{Appendix: Supplementary Methodology and Results}

\section{Dataset generation detail}
\label{app:dataset}

\paragraph{Generation procedure.}
Base problems are produced by 12 subcategory-specific template functions covering
arithmetic, percentages, rates, fractions, counting, geometry, compound interest,
and nested fractions across 2 to 4 hop depths. Each function draws numeric
parameters (prices, rates, counts, percentages) from fixed candidate sets using a
seeded \texttt{random.Random} instance, instantiates a natural-language question,
and computes each intermediate result exactly. Trace identities are 8-byte BLAKE2b
hashes of \texttt{(base\_problem\_id, variant)}, making each trace uniquely and
reproducibly addressable, and all hops downstream of an injection carry
\texttt{propagates\_prior\_error: true}.

\paragraph{Error-type motivation.}
The five error types are chosen to span distinct failure modes seen in real
multi-step reasoning:
\begin{itemize}
  \item \texttt{off\_by\_one} (correct value $+1$): small transcription or
    rounding errors that leave the answer superficially plausible. The hardest
    case for surface signals, since the corrupted value is coherent and close to
    correct.
  \item \texttt{wrong\_operator} (e.g.\ $\times \to +$): a conceptual mistake that
    combines the right quantities incorrectly. The result can be wildly off,
    testing whether probes respond to corruption magnitude rather than mere
    presence.
  \item \texttt{wrong\_unit} (e.g.\ hours instead of minutes, factor of 60):
    unit-tracking failures common in scientific and engineering chains, where the
    computation is correct but the semantic interpretation is wrong.
  \item \texttt{magnitude\_error} (correct value $\times 10$): a decimal-placement
    or order-of-magnitude mistake, structurally similar to the correct value but
    numerically large, testing sensitivity to scale.
  \item \texttt{wrong\_percentage\_base} ($100 - p$ instead of $p$): a systematic
    conceptual error confusing what is kept with what is removed. Common in
    financial and allocation chains, producing plausible-looking but wrong
    downstream values.
\end{itemize}

\section{Surface-signal definitions and behavioral flags}
\label{app:surface_signals}

\paragraph{Signal interpretation.}
The six signals in \Cref{eq:signals} test three distinct hypotheses about where
uncertainty might appear in the output distribution. Entropy-based signals
($s_\text{peak-ent}$, $s_\text{mean-ent}$, $s_\text{early-ent}$) measure
distributional spread across the full vocabulary; confidence signals
($s_\text{min-conf}$) measure top-token certainty; and $s_\text{vocab-gap}$
measures the margin between the top two candidates, which can remain large even
when entropy is moderate. Here $y_t$ denotes the token actually generated at
position $t$, so $s_\text{logprob}$ is a valid sequence log-probability under
sampling; under greedy decoding it reduces to the mean negative log-probability of
the top token, $-\tfrac{1}{G}\sum_t \log p_t^{(1)}$.

\paragraph{Behavioral flags.}
In addition to the scalar signals, four flags are extracted from the generated
text: presence of hedging language, presence of overconfident language, a
structured verbalized confidence score (where elicited), and an absurdity flag for
outputs containing logical impossibilities. These flags measure whether the
model's verbalized behavior reflects the internal state captured by the probe.

\section{Grouped cross-validation}
\label{app:grouped_cv}

To rule out leakage from shared base problems and templates, we re-run probe
evaluation under group-stratified $5$-fold cross-validation in which all variants
of a given base problem are confined to a single fold, so no template instance
appears in both train and test. \Cref{tab:grouped_cv} compares stratified and
grouped detection AUROC at the best layer and aggregation mode for each model.
Across all five models the two protocols agree to within $0.003$ AUROC, and on
three of the five the grouped estimate is marginally higher, so the near-perfect
detection reported in \Cref{tab:probe_summary} reflects a genuinely decodable
corruption state rather than memorised base-problem or template structure.

\begin{table*}[t]
\begin{center}
\small
\begin{tabular}{lccccc}
\toprule
Model & Layer & Mode & Stratified & Grouped & $\Delta$ \\
\midrule
Qwen2.5-3B        & 26 & last & 0.995 & $0.996 \pm 0.003$ & $+0.001$ \\
Llama-3.2-3B      & 11 & last & 0.995 & $0.992 \pm 0.006$ & $-0.003$ \\
Llama-3.1-8B      & 31 & mean & 0.997 & $0.998 \pm 0.005$ & $+0.001$ \\
Qwen3-4B (inst.)  & 20 & last & 0.991 & $0.994 \pm 0.005$ & $+0.003$ \\
Qwen3-4B (think)  & 20 & last & 0.991 & $0.994 \pm 0.005$ & $+0.003$ \\
\bottomrule
\end{tabular}
\end{center}
\caption{Best-layer detection AUROC under stratified versus base-problem-grouped
cross-validation. Both columns are computed at the layer and aggregation mode
listed, which is the best stratified configuration for that model in this run.
Grouped AUROC is reported as mean$\pm$std across the grouped folds.}
\label{tab:grouped_cv}
\end{table*}

\section{Held-out error-type generalisation}
\label{app:heldout_type}

To test whether the probe encodes an abstract corruption state rather than surface
signatures of individual error types, we train on four error types and evaluate
detection on the fifth, for all five leave-one-out splits. Clean traces are pooled
across splits; only the error-type composition of the training set varies.
\Cref{tab:heldout_type} reports detection AUROC on each held-out type at the best
layer for all five models. Every held-out type transfers above $0.91$, including
\texttt{off\_by\_one}, the hardest case on surface grounds, and \texttt{wrong\_unit}
and \texttt{magnitude\_error} transfer perfectly. A probe that never saw a given
error type during training still flags it at test time, so the probed direction is
not a per-type token detector.

\begin{table*}[t]
\begin{center}
\small
\setlength{\tabcolsep}{4pt}
\begin{tabular}{lccccc}
\toprule
Held-out type & Qwen2.5 & Llama-3.2 & Llama-3.1 & Qwen3 & Qwen3 \\
              & 3B      & 3B        & 8B        & inst. & think \\
\midrule
\texttt{off\_by\_one}      & .961 & .916 & .992 & .920 & .920 \\
\texttt{wrong\_operator}   & .991 & .930 & 1.00 & .999 & .999 \\
\texttt{wrong\_unit}       & 1.00 & 1.00 & 1.00 & 1.00 & 1.00 \\
\texttt{magnitude\_error}  & 1.00 & 1.00 & 1.00 & 1.00 & 1.00 \\
\texttt{wrong\_pct\_base}  & .996 & .993 & .987 & 1.00 & 1.00 \\
\midrule
Mean                       & .989 & .968 & .996 & .984 & .984 \\
\bottomrule
\end{tabular}
\end{center}
\caption{Leave-one-error-type-out detection AUROC at the best layer. Columns are
models; each row trains on the other four error types and tests on the named
held-out type. Leading zeros omitted; \texttt{wrong\_pct\_base} abbreviates
\texttt{wrong\_percentage\_base}.}
\label{tab:heldout_type}
\end{table*}
\section{Threshold calibration detail}
\label{app:thresholds}

The three thresholds in \Cref{eq:thresholds} are calibrated on out-of-fold scores
to avoid threshold overfitting while preserving all data for probe training. Here
$F^{-1}_{y=0}(0.99)$ is the 99th percentile of out-of-fold scores for clean
samples, giving $\tau_\text{conservative}$ a very low false-positive rate at the
cost of reduced recall, while $\tau_\text{loose} = 0.7\,\tau_\text{Youden}$ fires
more aggressively, rescuing more errors at higher broken counts. Only
$\tau_\text{loose} < \tau_\text{Youden}$ holds by construction; whether
$\tau_\text{conservative}$ sits above or below $\tau_\text{Youden}$ depends on the
clean and corrupted score distributions, and on the highly separable runs here the
two are nearly equal (see \Cref{fig:trajectory}). The firing rule of \Cref{eq:fire}
(fire when the probe score is at least $\tau$) then determines the firing-rate
ordering used in the sensitivity ablation (\Cref{sec:sensitivity}).

\section{Surface-signal and thinking-mode detail}
\label{app:surface_thinking}

\paragraph{Surface signals.}
\Cref{tab:surface} reports detection and failure-prediction AUROC for the
token-distribution signals. All are at or near chance for detection: peak entropy
is exactly $0.500$ for the instruct models shown, logprob uncertainty tops out at
$0.622$ (Llama-3.1-8B) and falls below $0.500$ for some Qwen variants, and none of
the signals shown exceeds $0.70$. Failure prediction is similarly uninformative,
with several
entries below chance (for example logprob uncertainty at $0.411$--$0.417$ on the
Llama models), indicating these signals are not merely uncorrelated with
propagation but mildly anti-predictive. The Qwen3-4B thinking model is the clearest
reversal: peak entropy drops to $0.308$ for detection, so tokens in error traces
are on average \emph{less} entropic than in clean traces, consistent with
thinking-mode suppression of surface uncertainty.

\begin{table}[h]
\begin{center}
\small
\setlength{\tabcolsep}{4pt}
\begin{tabular}{llcc}
\toprule
Model & Signal & Detect & Failure pred. \\
\midrule
\multirow{3}{*}{Llama-3.2-3B} & Logprob unc. & 0.574 & 0.417 \\
 & Peak entropy & 0.500 & 0.500 \\
 & Vocab gap & 0.451 & 0.585 \\
\midrule
\multirow{3}{*}{Llama-3.1-8B} & Logprob unc. & 0.622 & 0.411 \\
 & Peak entropy & 0.500 & 0.500 \\
 & Vocab gap & 0.432 & 0.621 \\
\midrule
\multirow{3}{*}{Qwen3-4B (think)} & Logprob unc. & 0.345 & 0.500 \\
 & Peak entropy & 0.308 & 0.500 \\
 & Vocab gap & 0.646 & 0.500 \\
\bottomrule
\end{tabular}
\end{center}
\caption{Representative surface-signal AUROC for detection and failure prediction
(three signals shown per model). All are at or near chance for detection, and
failure prediction is similarly uninformative.}
\label{tab:surface}
\end{table}

\paragraph{Thinking-mode ablation.}
\Cref{tab:thinking} contrasts the Qwen3-4B instruct and thinking variants. Probe
AUROC is essentially unchanged across the two modes (within $0.004$), while
final-answer accuracy degrades under thinking, from 29.1\% to 12.8\% on clean
problems and from 27.9\% to 1.2\% under error injection, despite roughly five
times more generated tokens. Because the variants share weights and differ only in
the inference-time template, this is evidence that the internal error encoding
persists even as the behavioral output channel degrades.

\begin{table}[h]
\begin{center}
\small
\begin{tabular}{lcc}
\toprule
Metric & Instruct & Thinking \\
\midrule
Best probe AUROC (last) & 0.990 & 0.989 \\
Best probe AUROC (mean) & 0.993 & 0.989 \\
Clean final correct (\%) & 29.1 & 12.8 \\
Error final correct (\%) & 27.9 & 1.2 \\
Hedging rate (\%) & 0.0 & 1.9 \\
Mean tokens (error-std) & 18.4 & 89.5 \\
\bottomrule
\end{tabular}
\end{center}
\caption{Thinking-mode ablation on Qwen3-4B. Probe AUROC is essentially unchanged
while final-answer accuracy degrades.}
\label{tab:thinking}
\end{table}

\section{Per-error-type intervention effects}
\label{app:per_error}

\Cref{tab:byerror} provides the per-error-type breakdown referenced in
\Cref{sec:intervention_results}. Each cell reports the change in accuracy (in
percentage points) relative to the baseline policy, together with the
corresponding rescued and broken counts in parentheses. Bold entries indicate the
best-performing policy for each error type.

For readability, several shortened names used in the main text correspond to the
implementation labels as follows: \texttt{wrong\_op} corresponds to
\texttt{wrong\_operator}; \texttt{magnitude} corresponds to
\texttt{magnitude\_error}; and \texttt{wrong\_pct\_base} corresponds to
\texttt{wrong\_percentage\_base}.

\begin{table*}[t]
\begin{center}
\small
\setlength{\tabcolsep}{8pt}
\begin{tabular}{lcrrr}
\toprule
Error type & $n$ & Reprompt & Replace prior & Branch + pick \\
\midrule
\texttt{wrong\_unit}      & 16 & $+0.0$ (0/0) & $\bm{+37.5}$ (7/1) & $+0.0$ (0/0) \\
\texttt{off\_by\_one}     & 13 & $+0.0$ (0/0) & $-7.7$ (4/5)       & $\bm{+15.4}$ (2/0) \\
\texttt{magnitude}        &  9 & $+0.0$ (0/0) & $\bm{+11.1}$ (3/2) & $\bm{+11.1}$ (1/0) \\
\texttt{wrong\_op}        & 16 & $+0.0$ (0/0) & $\bm{+6.2}$ (5/4)  & $\bm{+6.2}$ (1/0) \\
\texttt{wrong\_pct\_base} & 11 & $+0.0$ (0/0) & $-18.2$ (2/4)      & $-9.1$ (0/1) \\
\bottomrule
\end{tabular}
\end{center}
\caption{Per-error-type intervention effects on the Qwen2.5-3B intervention set
(65 traces across the five error types). Here $n$ denotes the number of examples
for each error type.}
\label{tab:byerror}
\end{table*}

\section{Pooling ablation}
\label{app:pooling}

In \Cref{sec:methods}, we report results using last-token activation pooling. For
completeness, we additionally trained probes using mean-pooled activations across
the full prompt. The peak performance is comparable between the two approaches:
on Qwen2.5-3B mean pooling achieves a maximum $\auroc$ of 0.986 at layer 27, while
last-token pooling achieves 0.991 at the same layer. The primary qualitative
difference appears in the earliest layers: under mean pooling the layer-0 probe
already achieves $\auroc \approx 0.91$, rather than remaining near chance, because
mean pooling aggregates information from tokens appearing after the corruption has
already been processed somewhere within the network; the per-token trajectories in
\Cref{app:trajectory} make this accumulation explicit.

Both aggregation modes are trained at every layer and the stronger is selected per
layer (\Cref{sec:stage2}); the headline figures in \Cref{tab:probe_summary} use
whichever mode is stronger per model (mean pooling for Llama-3.1-8B, last-token
elsewhere). Last-token pooling preserves a cleaner interpretation of information
flow, since it measures what the model has computed at a specific point in the
sequence, so we adopt it as the default for the information-flow narrative. One
particularly striking feature of the last-token profile is the sharp increase
from layer 0 to layer 1, where a single transformer block contributes
approximately 0.39 AUROC. This behavior is obscured under mean pooling, since the
layer 0 representation is already highly informative due to prompt-level
aggregation. Whether this jump is driven by a specific attention head detecting
token inconsistencies or by an MLP-based feature remains an open mechanistic
question. We identify this as a promising direction for future investigation.

\section{Token-position probe trajectories}
\label{app:trajectory}

\begin{figure*}[t]
\centering
\includegraphics[width=\textwidth]{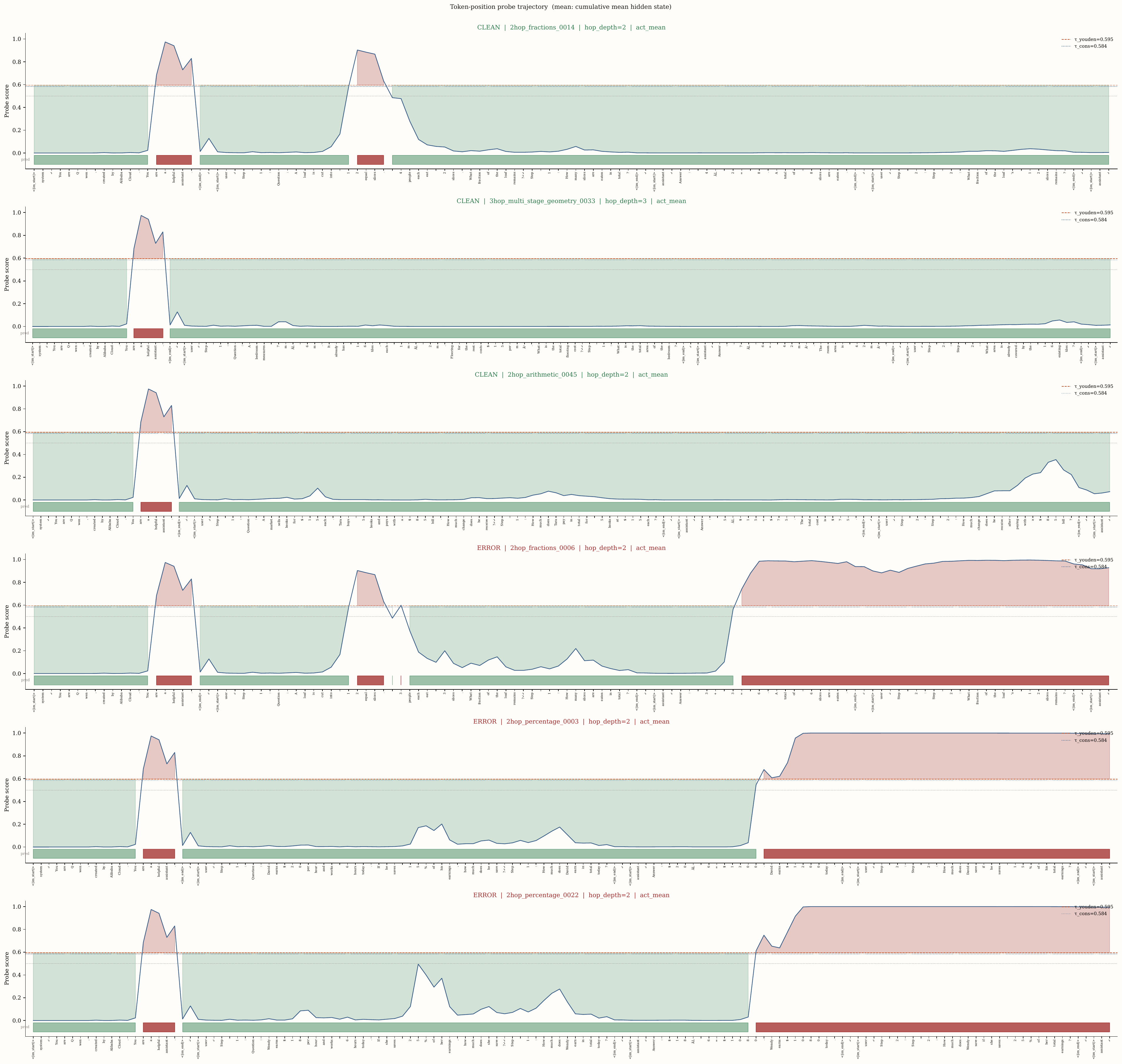}
\caption{Per-token probe-score trajectories under cumulative-mean pooling for
three \texttt{CLEAN} (top) and three \texttt{ERROR} (bottom) Qwen traces.
The horizontal axis is generated token position and the vertical axis is the probe
score; dashed lines mark the Youden and conservative thresholds. Error trajectories
move above threshold once the corrupted operand enters the running mean and stay
there, whereas clean trajectories settle below after any early excursion,
illustrating why mean pooling is informative even at shallow layers
(\Cref{app:pooling}).}
\label{fig:trajectory}
\end{figure*}

To make the cumulative-mean mechanism concrete, \Cref{fig:trajectory} traces the
probe score \emph{position by position} as the running mean of hidden states grows
during generation. Formally, at generated position $t$ the probe is applied to the
prefix mean
$\bar{\mathbf{a}}^{(\ell)}_t = \tfrac{1}{t}\sum_{s=1}^{t}\mathbf{h}^{(\ell)}_s$,
so the value at the final position equals the single mean-pooled score analysed in
\Cref{app:pooling}. The figure therefore decomposes that scalar into the
trajectory that produces it. Three \texttt{CLEAN} traces (top) and three
\texttt{ERROR} traces (bottom) are shown for Qwen at the best mean-pooling layer,
with the Youden and conservative firing thresholds ($\tau_\text{Youden}=0.595$,
$\tau_\text{cons}=0.584$) marked. The two thresholds nearly coincide here, an
instance of the empirical near-equality noted in \Cref{app:thresholds} rather than
a guaranteed ordering.

Two patterns are visible. First, clean and corrupted traces separate at the level
that matters for pooling: although a clean trajectory may briefly excurse above
threshold at an early position, its \emph{persistent} cumulative-mean score
settles below both thresholds, whereas error trajectories stay above once the
corrupted operand enters the running mean, reproducing at the single-trace level
the
near-perfect separability reported in aggregate (\Cref{tab:probe_summary}).
Second, and more informative for the pooling comparison, the error trajectories
cross the threshold \emph{once the corrupted operand enters the running mean} and
remain there for the rest of the sequence, rather than spiking only at the final
token. Because every subsequent position averages in the already-corrupted prefix,
the mean-pooled score is driven above threshold well before the end of generation.
This is the mechanism behind the otherwise surprising result that mean pooling
attains $\auroc \approx 0.91$ at layer~0 (\Cref{app:pooling}): the corruption
signal is smeared across positions by the averaging operation rather than
localised at the final token, which is precisely the property that last-token
pooling avoids and the reason we adopt last-token pooling as the primary
aggregation in the main text. The selected error traces span an arithmetic
operator substitution (\texttt{2hop\_fractions\_0006}, $3{+}3$ in place of
$3{\times}3$) and two magnitude errors in percentage chains
(\texttt{2hop\_percentage\_0003}, \texttt{2hop\_percentage\_0022}), indicating the
behaviour is not specific to a single error type.

\section{Sensitivity ablation detail}
\label{app:sensitivity}

The branch-and-pick factorial (\Cref{sec:sensitivity}) crosses $K \in \{1,2,4,8\}$
with the three threshold modes. When $K \neq 4$, candidates are drawn from the
same temperature schedule: the first $K$ entries of $(0.7, 0.85, 1.0, 1.15)$ for
$K < 4$, and each of the four temperatures sampled twice for $K = 8$. The $K$
ablation therefore varies branch count while holding the temperature range fixed.
Conservative thresholds fire rarely, reducing the risk of breaking correct traces
but missing some errors, whereas loose thresholds fire aggressively, rescuing more
errors at higher broken counts. The interaction with $K$ reveals whether more
candidates compensate for a noisier trigger.

\section{Persistence versus peak analysis}
\label{app:persistence}

For each multi-hop error trace we compute two scalars, namely \emph{peak}
(maximum probe score across hops) and \emph{persistence} (fraction of
hops exceeding $\tau_\text{Youden}$). $\text{AUROC}_\text{fail}$ is
computed for each scalar, and neither scalar separates degraded from recovered
traces in any model, refuting the pre-registered persistence-beats-peak
hypothesis.

\end{document}